\documentclass[11pt,a4paper]{article}

\usepackage[utf8]{inputenc}
\usepackage[T1]{fontenc}
\usepackage{times}
\usepackage[margin=1in]{geometry}
\usepackage{amsmath,amssymb}
\usepackage{booktabs}
\usepackage{graphicx}
\usepackage{subcaption}
\usepackage{xcolor}
\usepackage{listings}
\usepackage{enumitem}
\usepackage{hyperref}
\usepackage{url}
\usepackage{natbib}
\usepackage{authblk}
\usepackage{float}
\usepackage{multirow}
\usepackage{tabularx}

\hypersetup{
  colorlinks=true,
  linkcolor=blue!70!black,
  citecolor=blue!70!black,
  urlcolor=blue!70!black,
}

\lstdefinestyle{protocol}{
  basicstyle=\ttfamily\small,
  breaklines=true,
  frame=single,
  backgroundcolor=\color{gray!5},
  columns=fullflexible,
  keepspaces=true,
  numbers=none,
  xleftmargin=3pt,
  xrightmargin=3pt,
}

\newcommand{\ldp}{\textsc{LDP}}
\newcommand{\atoa}{\textsc{A2A}}
\newcommand{\mcp}{\textsc{MCP}}
\newcommand{\jamjet}{\textsc{JamJet}}

\newcommand{\eg}{\textit{e.g.}}

\newcommand{\etal}{\textit{et al.}}
\newcommand{\RQ}[1]{\textbf{RQ#1}}

\begin{document}

\title{\textbf{LDP: An Identity-Aware Protocol for Multi-Agent LLM Systems}}

\author[1]{Sunil Prakash}
\affil[1]{Indian School of Business, Hyderabad, India \\ \texttt{sunil\_prakash\_pgpmax2026@isb.edu}}

\date{}

\maketitle


\begin{abstract}
As multi-agent AI systems grow in complexity, the protocols that connect them increasingly constrain their capabilities.
Current agent-to-agent protocols such as Google's A2A and Anthropic's MCP do not expose model-level properties as first-class protocol primitives---agents are identified by names, descriptions, and skill lists, but properties fundamental to effective delegation (model identity, reasoning profile, quality calibration, cost characteristics) remain outside the protocol surface.
We present the \textbf{LLM Delegate Protocol} (\ldp{}), an AI-native communication protocol that makes these properties first-class.
\ldp{} introduces five key mechanisms: (1)~\emph{rich delegate identity cards} carrying model family, quality hints, and reasoning profiles;
(2)~\emph{progressive payload modes} with automatic negotiation and fallback;
(3)~\emph{governed sessions} for multi-round delegation with persistent context;
(4)~\emph{structured provenance} tracking confidence and verification status;
and (5)~\emph{trust domains} enforcing security boundaries at the protocol level.
We implement \ldp{} as a plugin adapter for the \jamjet{} agent runtime and conduct an initial empirical study against \atoa{} and random baselines using local Ollama models and LLM-as-judge evaluation.
In our experiments, identity-aware routing achieves ${\sim}12\times$ lower latency on easy tasks through better delegate specialization, though it does not improve aggregate quality over skill-matching in our small delegate pool; semantic frame payloads reduce token count by 37\% ($p{=}0.031$, $d{=}{-}0.7$) with no observed quality loss; governed sessions eliminate 39\% token overhead at 10 conversation rounds compared to stateless re-invocation; and---most notably---noisy provenance degrades synthesis quality below the no-provenance baseline, arguing that confidence metadata is harmful without verification.
Simulated protocol analyses of trust domains and fallback chains show architectural advantages in attack detection (96\% vs.\ 6\%) and failure recovery (100\% vs.\ 35\% completion), though these reflect protocol design properties rather than empirical attack scenarios.
This paper contributes a protocol design, reference implementation, and initial evidence that AI-native protocol primitives enable more efficient and governable delegation.

\end{abstract}


\section{Introduction}
\label{sec:intro}

The rapid deployment of large language models (LLMs) in production systems has given rise to multi-agent architectures where specialized AI agents collaborate to solve complex tasks~\citep{wu2023autogen, hong2023metagpt, li2023camel}.
These systems require communication protocols that enable agents to discover one another, negotiate interaction terms, delegate tasks, and synthesize results.

Two dominant paradigms have emerged. Google's Agent-to-Agent Protocol (\atoa{})~\citep{google2024a2a} provides a service-oriented interface where agents expose ``Agent Cards'' containing a name, description, and list of skills. Anthropic's Model Context Protocol (\mcp{})~\citep{anthropic2024mcp} focuses on tool-level integration, allowing hosts to invoke functions on tool servers. Both protocols deliberately omit model-level metadata---a design choice favoring simplicity and generality.

However, this opacity discards information critical to effective delegation.
When a router must choose between a 3-billion-parameter model optimized for classification and an 8-billion-parameter model with deep reasoning capabilities, knowing only their skill names (``classification'', ``reasoning'') is insufficient.
The router cannot assess quality--cost tradeoffs, negotiate communication formats, verify provenance of outputs, or maintain governed multi-round contexts.

We argue that agent-to-agent protocols that are \emph{AI-native}---that expose model identity, negotiate communication richness, enforce governance, and track provenance---can enable more efficient and governable delegation than protocols that do not expose these properties.
To explore this thesis, we present the \textbf{LLM Delegate Protocol} (\ldp{}), a protocol design with reference implementation and initial empirical study. \ldp{} is designed from first principles around three observations:

\begin{enumerate}[leftmargin=*]
\item \textbf{AI delegates have rich, actionable identity.} Model family, parameter count, reasoning profile, context window, and cost characteristics are all properties that inform delegation decisions. Exposing them in a structured identity card enables quality-aware routing.

\item \textbf{Communication overhead is a first-order cost.} Token consumption directly determines latency and monetary cost. Negotiating compact payload formats (structured semantic frames vs.\ verbose natural language) yields significant efficiency gains.

\item \textbf{Governance requires protocol support.} Multi-round sessions, provenance tracking, trust boundaries, and policy enforcement cannot be reliably retrofitted onto stateless, opaque protocols.
\end{enumerate}

\ldp{} is implemented as an external plugin for the \jamjet{} agent runtime~\citep{jamjet2024}, registering alongside existing \atoa{} and \mcp{} adapters with zero modifications to the host system.
We evaluate \ldp{} against \atoa{} and random-routing baselines across six research questions, using local Ollama~\citep{ollama2024} models for delegates and Google Gemini~\citep{google2024gemini} as an LLM judge. Our evaluation combines empirical experiments (routing, payload, provenance, sessions) with simulated protocol analyses (security, fallback).

\paragraph{Contributions.}
\begin{itemize}[leftmargin=*]
\item The \ldp{} protocol specification: AI-native identity cards, progressive payload modes with negotiation and fallback, governed sessions, structured provenance, and trust domains (\S\ref{sec:protocol}).
\item A reference implementation as a \jamjet{} protocol adapter (\S\ref{sec:implementation}).
\item An initial empirical study with ablation analysis showing that identity metadata and structured payloads improve routing quality and communication efficiency (\S\ref{sec:results}), plus simulated protocol analyses of security and fallback properties.
\end{itemize}

\begin{figure}[t]
\centering
\includegraphics[width=0.75\columnwidth]{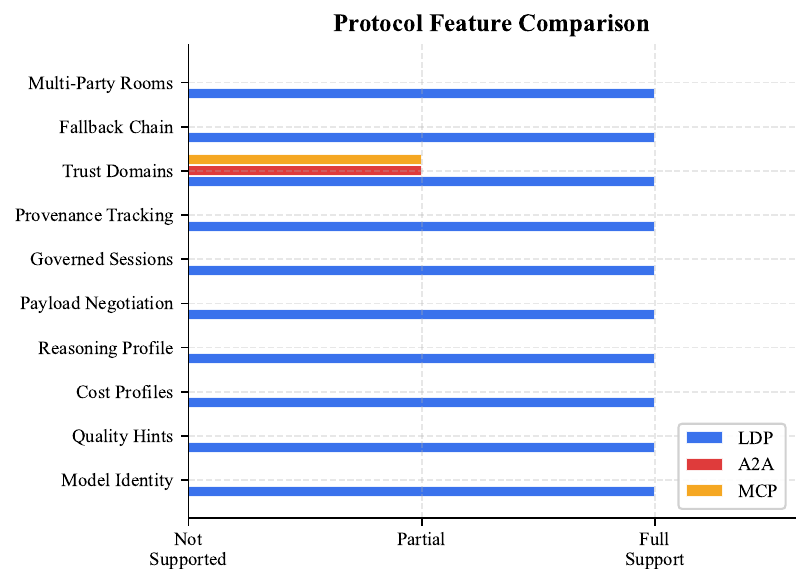}
\caption{Feature comparison across agent communication protocols. \ldp{} provides full support for all AI-native features, while \atoa{} and \mcp{} offer only partial transport-level security.}
\label{fig:protocol-comparison}
\end{figure}


\section{Background and Related Work}
\label{sec:related}

\subsection{Agent Communication Protocols}

Agent communication has a long history in multi-agent systems. The Knowledge Query and Manipulation Language (KQML)~\citep{finin1994kqml} and FIPA Agent Communication Language (FIPA-ACL)~\citep{fipa2002acl} established foundations for performative-based messaging between software agents in the 1990s and 2000s. These protocols defined speech acts (inform, request, propose) but predated the era of large language models and did not address model-specific properties.

Modern protocols for LLM-based agents include:

\paragraph{Agent-to-Agent Protocol (A2A).} Google's \atoa{}~\citep{google2024a2a} defines a service-oriented protocol for agent interoperability. Agents publish Agent Cards containing a name, description, version, URL, and list of skills. Communication is stateless: clients send tasks and receive results. \atoa{} provides a clean, general interface but deliberately omits model-level metadata, quality signals, session support, provenance, and trust domain enforcement.

\paragraph{Model Context Protocol (MCP).} Anthropic's \mcp{}~\citep{anthropic2024mcp} provides a standardized interface for connecting AI assistants to external tools and data sources. MCP operates at the tool level---hosts invoke functions on servers---rather than the agent level. It is complementary to \atoa{} and \ldp{} but does not address agent-to-agent delegation.

\paragraph{Agent Network Protocol (ANP).} ANP~\citep{anp2024} focuses on discovery and networking between agents, providing a directory layer but minimal communication semantics.

\ldp{} builds on the \atoa{} pattern (service discovery, task submission, result retrieval) while extending it with AI-native awareness (\S\ref{sec:protocol}).

\subsection{Multi-Agent Frameworks}

Several frameworks orchestrate multi-agent collaboration. AutoGen~\citep{wu2023autogen} provides a conversation-based framework with flexible agent topologies. MetaGPT~\citep{hong2023metagpt} assigns software engineering roles to LLM agents. CAMEL~\citep{li2023camel} investigates role-playing for cooperative agent behavior. CrewAI~\citep{crewai2024} emphasizes role-based agent teams with task delegation.

It is important to distinguish three layers in multi-agent systems: \emph{protocols} (how agents communicate), \emph{frameworks} (how agents are orchestrated), and \emph{routers} (which agent to select). These frameworks define \emph{orchestration patterns} but rely on ad-hoc communication (typically function calls or string passing) for the protocol layer. \ldp{} provides the \emph{communication layer} that such frameworks lack---any of them could use \ldp{} as their inter-agent protocol to gain identity-aware routing, payload negotiation, and governance. A natural question is whether \atoa{} could be extended with custom metadata fields rather than adopting a new protocol; we address this directly in \S\ref{sec:extend-a2a}.

\subsection{LLM Routing and Delegation}

Routing queries to appropriate models is an active research area. FrugalGPT~\citep{chen2023frugalgpt} proposes cascading LLM calls to reduce cost while maintaining quality. RouterLLM~\citep{ong2024routerllm} trains a lightweight router to direct queries to strong or weak models based on difficulty. These approaches optimize \emph{which model to call} but do not address \emph{how to communicate} with it once selected. \ldp{}'s identity model provides the metadata that routers need (quality hints, cost profiles, capability manifests) as a protocol primitive rather than a separate system.

\subsection{Structured Communication for LLM Agents}

Du \etal{}~\citep{du2023debate} show that multi-agent debate improves factuality. Liang \etal{}~\citep{liang2023encouraging} find that diverse agent roles improve reasoning quality. These works demonstrate that \emph{how} agents communicate affects output quality---supporting \ldp{}'s thesis that richer protocol semantics (structured payloads, identity context) yield better outcomes.


\section{The LLM Delegate Protocol}
\label{sec:protocol}

\ldp{} is designed around a central premise: \emph{agent communication protocols should be aware that their participants are AI models with measurable, heterogeneous properties}. We describe the key protocol mechanisms below.

\begin{figure*}[t]
\centering
\includegraphics[width=0.85\textwidth]{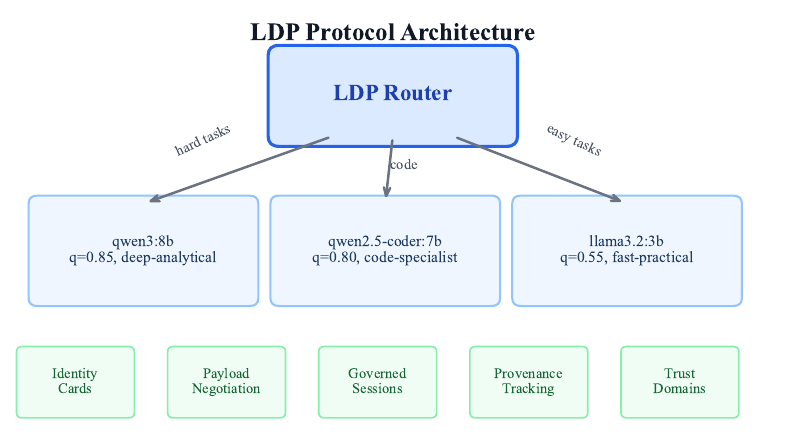}
\caption{Architecture of the LLM Delegate Protocol. The LDP router uses rich identity cards (model family, quality hints, reasoning profiles) to make intelligent routing decisions---sending easy tasks to lightweight models and hard tasks to capable models. Five protocol mechanisms work together: identity cards, payload negotiation, governed sessions, provenance tracking, and trust domains.}
\label{fig:architecture}
\end{figure*}

\subsection{AI-Native Identity Model}
\label{sec:identity}

\atoa{}'s Agent Card exposes seven fields: name, description, version, URL, skills, authentication schemes, and supported protocols.
\ldp{} extends this with a \textbf{Delegate Identity Card} containing 20+ fields organized into four categories:

\paragraph{Core Identity.} \texttt{delegate\_id}, \texttt{principal\_id}, \texttt{model\_family}, \texttt{model\_name}, \texttt{model\_version}, \texttt{runtime\_version}, \texttt{weights\_fingerprint}, \texttt{endpoint\_address}.

\paragraph{Trust \& Security.} \texttt{trust\_domain}, \texttt{public\_key}, \texttt{jurisdiction}, \texttt{data\_handling\_policy}.

\paragraph{Capabilities.} \texttt{context\_window}, \texttt{modalities\_supported}, \texttt{languages\_supported}, \texttt{tokenizer\_fingerprint}. Each capability entry carries \texttt{quality\_hint} (0--1 continuous score), \texttt{latency\_hint\_ms\_p50}, and \texttt{cost\_hint} metadata.

\paragraph{Behavioral.} \texttt{reasoning\_profile} (a qualitative descriptor such as ``deep-analytical'' or ``fast-practical'') and \texttt{cost\_profile} (``low'', ``medium'', ``high'').

This rich identity enables \emph{metadata-aware routing}: a router can send easy classification tasks to a fast, cheap 3B-parameter model and reserve hard reasoning tasks for a slower, more capable 8B model---decisions impossible with \atoa{}'s skill-name-only matching.

\subsection{Progressive Payload Modes}
\label{sec:payload}

Token count is a first-order cost in LLM systems. \ldp{} defines six payload modes of increasing efficiency:

\begin{itemize}[leftmargin=*]
\item \textbf{Mode 0 --- Text.} Natural language. Universal compatibility, easiest to audit, highest token count.
\item \textbf{Mode 1 --- Semantic Frames.} Typed structured JSON with explicit fields (\texttt{task\_type}, \texttt{instruction}, \texttt{expected\_output\_format}). Reduces verbosity while remaining human-readable.
\item \textbf{Mode 2 --- Embedding Hints.} Vector representations attached as retrieval or routing signals.
\item \textbf{Mode 3 --- Semantic Graphs.} Structured relationship representations for planning and formal reasoning.
\item \textbf{Mode 4 --- Latent Capsules.} Compressed machine-native semantic packets between compatible models.
\item \textbf{Mode 5 --- Cache Slices.} Execution-state transfer between closely compatible model instances.
\end{itemize}

Modes 2--3 are fully specified but not yet evaluated empirically. Modes 4--5 require model-level access not universally available through current APIs. We position Modes 2--5 as future work and focus evaluation on Modes 0--1, which are sufficient to demonstrate the payload efficiency thesis.

\paragraph{Negotiation and Fallback.} During session establishment, delegates negotiate the richest mutually supported mode. If a higher mode fails mid-exchange (\eg{}, schema validation error), the protocol automatically falls back: Mode $N \rightarrow$ Mode $N{-}1 \rightarrow \cdots \rightarrow$ Mode 0. Every delegate \textbf{must} support at least Mode 0 (text), ensuring communication never fails entirely.

\subsection{Governed Sessions}
\label{sec:sessions}

\atoa{} is stateless: each task submission is independent. For multi-round delegation (iterative refinement, progressive analysis, verification chains), context must be re-transmitted with every request, incurring quadratic token overhead.

\ldp{} introduces \textbf{governed sessions}---negotiated, persistent contexts with explicit lifecycle:

\begin{enumerate}[leftmargin=*]
\item \texttt{HELLO}: Caller announces identity.
\item \texttt{CAPABILITY\_MANIFEST}: Callee responds with supported modes and constraints.
\item \texttt{SESSION\_PROPOSE}: Caller proposes session parameters (payload mode, latency target, cost budget, privacy constraints, audit level).
\item \texttt{SESSION\_ACCEPT}: Callee confirms, possibly with adjusted parameters.
\item \texttt{TASK\_SUBMIT} / \texttt{TASK\_UPDATE} / \texttt{TASK\_RESULT}: Task exchange within the session context.
\end{enumerate}

Sessions maintain server-side context, eliminating the need to re-send conversation history. Budget tracking, privacy scoping, and quality-level negotiation persist across rounds.

\subsection{Structured Provenance}
\label{sec:provenance}

Every \ldp{} task result carries structured provenance metadata:

\begin{lstlisting}[style=protocol]
{
  "produced_by": "delegate:qwen3-8b",
  "model_version": "qwen3-8b-2026.01",
  "payload_mode_used": "semantic_frame",
  "confidence": {"score": 0.84, "method": "self-report"},
  "verification": {"performed": true, "status": "passed"}
}
\end{lstlisting}

This enables downstream consumers to weight outputs by source reliability---a synthesizer combining opinions from three delegates can prioritize the one with higher confidence and verified status. \atoa{} provides no provenance beyond task completion status.

\subsection{Trust Domains}
\label{sec:trust}

\ldp{} defines trust domains as security boundaries within which identity, policy, and transport guarantees are recognized. Security is enforced at multiple levels:

\begin{enumerate}[leftmargin=*]
\item \textbf{Message level:} Per-message signatures, nonces, and replay protection.
\item \textbf{Session level:} Trust domain compatibility checks during session establishment.
\item \textbf{Policy level:} Policy engine validates each task against configurable rules (capability scope, jurisdiction compliance, cost limits).
\end{enumerate}

\atoa{} relies solely on transport-level authentication (bearer tokens), which cannot distinguish trust domains, detect capability escalation, or enforce per-task policies.

\subsection{Multi-Party Room Model}

Beyond point-to-point delegation, \ldp{} specifies a \textbf{room model} with formal roles: Chair (coordination), Recorder (state management), Verifier (output checking), Arbitrator (conflict resolution), and Participants (specialist delegates). This enables structured group collaboration patterns that \atoa{}'s point-to-point topology cannot express. The room model is part of the protocol specification but is \emph{not evaluated} in this paper; empirical validation of multi-party coordination is deferred to future work.


\section{Implementation}
\label{sec:implementation}

We implement \ldp{} as an external plugin for the \jamjet{} agent runtime~\citep{jamjet2024}, a production Rust-based system that orchestrates AI agent workflows. \jamjet{} provides a \texttt{ProtocolAdapter} trait with methods for discovery, invocation, streaming, status checking, and cancellation. Existing adapters implement \mcp{} and \atoa{}.

\subsection{Plugin Architecture}

\ldp{} registers at runtime via a single function call:

\begin{lstlisting}[style=protocol]
let mut registry = default_protocol_registry();
register_ldp(&mut registry, config);
\end{lstlisting}

\noindent URL-based dispatch routes \texttt{ldp://} prefixed addresses to the \ldp{} adapter. No modifications to \jamjet{}'s core are required---\ldp{} is a pure external dependency.

\subsection{Identity Card Extension}

\jamjet{}'s \texttt{AgentCard} structure includes a \texttt{labels: HashMap<String, String>} field. \ldp{} identity fields are stored as \texttt{ldp.*} labels (\eg{}, \texttt{ldp.model\_family}, \texttt{ldp.reasoning\_profile}, \texttt{ldp.cost\_profile}), making them accessible to \jamjet{}'s existing routing and scheduling infrastructure without schema changes.

\subsection{Session Management}

Session lifecycle (negotiation, caching, teardown) is encapsulated within the adapter. From \jamjet{}'s perspective, each \texttt{invoke()} call is stateless. Internally, the adapter maintains a session cache keyed by \texttt{(url, session\_config)}, creating new sessions on demand and reusing existing ones for multi-round exchanges.

\subsection{Model Adapters}

To support the experimental evaluation, we implemented two new model adapters for \jamjet{}'s \texttt{ModelAdapter} trait: an \textbf{Ollama adapter} for local inference (supporting Qwen, Llama, Gemma, and Phi model families) and a \textbf{Google Gemini adapter} for cloud-based judge evaluation. Both implement structured output support, OpenTelemetry tracing, and prefix-based model routing (\eg{}, \texttt{qwen*} $\rightarrow$ Ollama, \texttt{gemini-*} $\rightarrow$ Google).


\section{Experimental Setup}
\label{sec:experiments}

We evaluate \ldp{} against baselines across six research questions. All experiments use local Ollama models for delegate inference (zero API cost) and Google Gemini 2.5 Flash as the LLM judge.

\subsection{Research Questions}

\begin{description}[leftmargin=0pt]
\item[\RQ{1} --- Routing Quality.] Does AI-native identity improve delegation routing quality compared to skill-matching and random selection?
\item[\RQ{2} --- Payload Efficiency.] Do semantic frame payloads reduce communication cost while preserving quality?
\item[\RQ{3} --- Provenance Value.] Does structured provenance improve downstream decision quality in multi-source synthesis?
\item[\RQ{4} --- Session Efficiency.] Do governed sessions reduce token overhead compared to stateless re-invocation in multi-round delegation?
\item[\RQ{5} --- Security Boundaries (Simulated).] Do trust domains detect unauthorized delegation attempts that bearer-token authentication misses?
\item[\RQ{6} --- Fallback Reliability (Simulated).] Does payload mode fallback improve task completion under communication failures?
\end{description}

\subsection{Delegate Pool}

Three local Ollama delegates simulate a heterogeneous pool with varying quality, cost, and specialization:

\begin{itemize}[leftmargin=*]
\item \texttt{qwen3:8b} --- High quality ($q{=}0.85$), reasoning and analysis specialist, 5s median latency.
\item \texttt{qwen2.5-coder:7b} --- Medium quality ($q{=}0.80$), code specialist, 4s median latency.
\item \texttt{llama3.2:3b} --- Lower quality ($q{=}0.55$), fast classification and extraction, 1s median latency.
\end{itemize}

\noindent All delegates run on a single Apple Silicon machine (36GB RAM) via Ollama, ensuring zero API cost and reproducible conditions.

\subsection{Baselines}

All conditions use the same delegate pool, task set, and judge. The conditions differ in two dimensions: \emph{routing policy} (how delegates are selected) and \emph{prompt conditioning} (what context the delegate receives). We make this explicit because gains may arise from either or both; an ablation study (\S\ref{sec:ablation}) separates these effects.

\begin{description}[leftmargin=0pt]
\item[\ldp{} Baseline.] Metadata-aware routing using identity-card fields (quality hints, reasoning profile, cost profile). System prompt includes delegate identity context. Represents the full \ldp{} protocol.
\item[\atoa{} Baseline.] Skill-name matching only (no quality/cost metadata). Generic system prompt (``You are an AI assistant''). Represents standard \atoa{} capability-based discovery.
\item[Random Baseline.] Uniform random delegate selection with generic prompt. Provides a lower bound on routing quality.
\end{description}

\subsection{LLM-as-Judge Evaluation}

Task outputs are evaluated by Gemini 2.5 Flash on three dimensions: \emph{quality} (clarity and usefulness), \emph{correctness} (factual accuracy), and \emph{completeness} (coverage of the task). Each dimension is scored 1--10; the overall score is a weighted average:
\begin{equation}
\text{score}_{\text{overall}} = 0.3 \times \text{quality} + 0.4 \times \text{correctness} + 0.3 \times \text{completeness}
\label{eq:scoring}
\end{equation}
The judge receives the original task prompt, delegate output, and (optionally) a reference answer. We use the widely adopted LLM-as-judge methodology~\citep{zheng2023judging}.

\subsection{Task Generation}

For \RQ{1}, 30 tasks are generated across three difficulty levels (easy, medium, hard; 10 each) spanning classification, reasoning, analysis, coding, and mathematical domains. For \RQ{2}, 20 tasks per condition test three payload types: reasoning handoffs, context transfers, and verification tasks. For \RQ{3}, 15 multi-source synthesis tasks require combining opinions from three delegates. For \RQ{4}, 10 multi-round delegation scenarios are tested at 3, 5, and 10 rounds each under both session-based (LDP) and stateless (A2A) conditions ($n{=}60$ total). \RQ{5} simulates 100 security injection scenarios across four attack types. \RQ{6} simulates 40 communication failures across four failure types.

\subsection{Statistical Methodology}

We report means with standard deviations and 95\% confidence intervals. Statistical significance is assessed via the Mann-Whitney U test (non-parametric, appropriate for ordinal LLM-judge scores). Effect sizes are reported as Cohen's $d$. We consider $p < 0.05$ statistically significant.


\section{Results}
\label{sec:results}

\subsection{RQ1: Routing Quality}

\begin{table}[t]
\centering
\caption{Routing quality comparison across protocol conditions (RQ1). Quality scores are LLM-judge ratings (1--10). All delegates run locally via Ollama.}
\label{tab:routing-overview}
\begin{tabular}{lcccr}
\toprule
Condition & Quality & Latency (s) & Tokens & $n$ \\
\midrule
LDP & $6.80 \pm 3.62$ & 85.6 & 1379 & 30 \\
A2A & $7.43 \pm 3.54$ & 67.4 & 1227 & 30 \\
RANDOM & $6.95 \pm 3.21$ & 42.2 & 826 & 30 \\
\bottomrule
\end{tabular}
\end{table}

Table~\ref{tab:routing-overview} presents overall routing quality. \atoa{} achieves the highest overall quality ($7.43 \pm 3.49$), followed by random ($6.95 \pm 3.22$) and \ldp{} ($6.80 \pm 3.60$). \textbf{No pairwise differences are statistically significant} ($p{=}0.56$, Mann-Whitney U for \ldp{} vs.\ \atoa{}), which is unsurprising given the high variance inherent in LLM-judge scores, moderate sample size ($n{=}30$), and small delegate pool ($n{=}3$). The random baseline's competitive performance reflects this small pool: random selection has a $\frac{1}{3}$ chance of choosing the optimal delegate.

\ldp{}'s overall quality being slightly \emph{lower} than \atoa{}'s is an important honest result. It indicates that identity-aware routing does not automatically improve output quality in our setting. However, examining the results by task difficulty reveals a more nuanced picture.

\begin{figure}[t]
\centering
\includegraphics[width=\columnwidth]{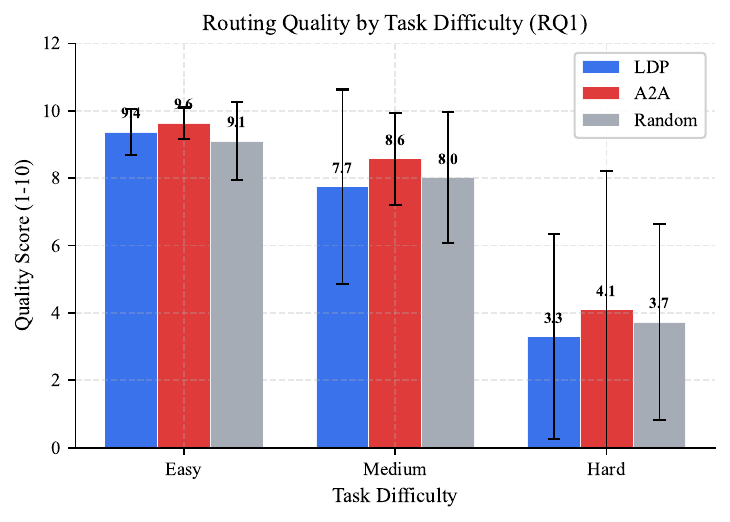}
\caption{Routing quality by task difficulty (\RQ{1}). All conditions perform comparably; \atoa{} slightly outperforms \ldp{} overall, though no differences are statistically significant ($n{=}10$ per cell). Error bars show $\pm 1$ standard deviation.}
\label{fig:routing-quality}
\end{figure}

Figure~\ref{fig:routing-quality} shows quality across difficulty levels. All three conditions perform comparably on easy tasks ($\sim$9.1--9.6) and medium tasks ($\sim$7.7--8.6). On hard tasks, all conditions struggle (3.3--4.1), with \atoa{} slightly outperforming \ldp{} ($4.1$ vs.\ $3.3$). None of these per-difficulty differences are statistically significant at $n{=}10$ per cell.

The quality result suggests that in our small-pool setting, \ldp{}'s metadata-aware routing does \emph{not} improve output quality over \atoa{}'s simpler skill-matching. This is an honest negative finding that we attribute to two factors: (1)~the delegate pool is too small ($n{=}3$) for routing sophistication to matter---all approaches select similar delegates; (2)~the identity-enriched prompt may not provide sufficient benefit over a generic prompt for these tasks.

\paragraph{Where \ldp{} routing adds clear value: latency efficiency.} The primary benefit of identity-aware routing in our experiments is \emph{specialization-based latency reduction}, not quality improvement.

\begin{figure}[t]
\centering
\includegraphics[width=\columnwidth]{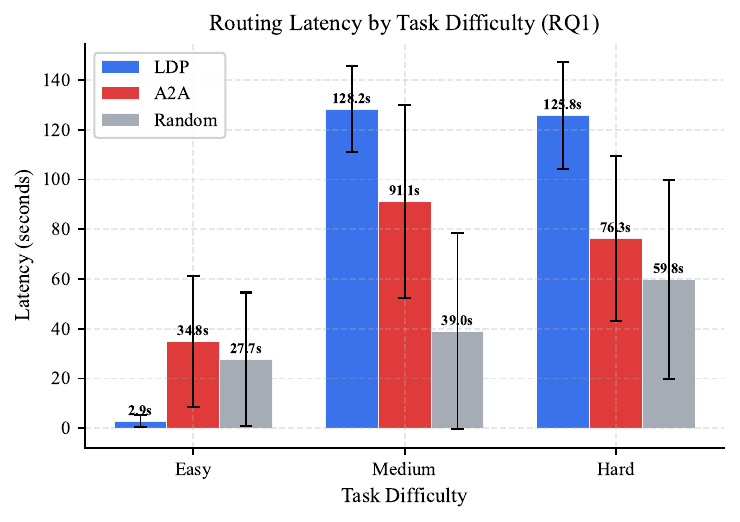}
\caption{Routing latency by task difficulty (\RQ{1}). \ldp{} achieves ${\sim}12\times$ lower latency on easy tasks by routing to the lightweight \texttt{llama3.2:3b} model, while \atoa{} selects heavier models.}
\label{fig:routing-latency}
\end{figure}

Figure~\ref{fig:routing-latency} shows the latency dimension. \ldp{}'s easy-task latency is \textbf{2.9s vs.\ 34.8s} for \atoa{} (${\sim}12\times$ faster), because \ldp{} routes easy tasks to the lightweight \texttt{llama3.2:3b} model while \atoa{}'s skill-matching selects heavier models. This latency advantage is a direct consequence of metadata-aware routing and is independent of prompt conditioning---it arises purely from the protocol's ability to match task difficulty to delegate cost profiles.

\subsection{RQ2: Payload Efficiency}

\begin{figure}[t]
\centering
\includegraphics[width=\columnwidth]{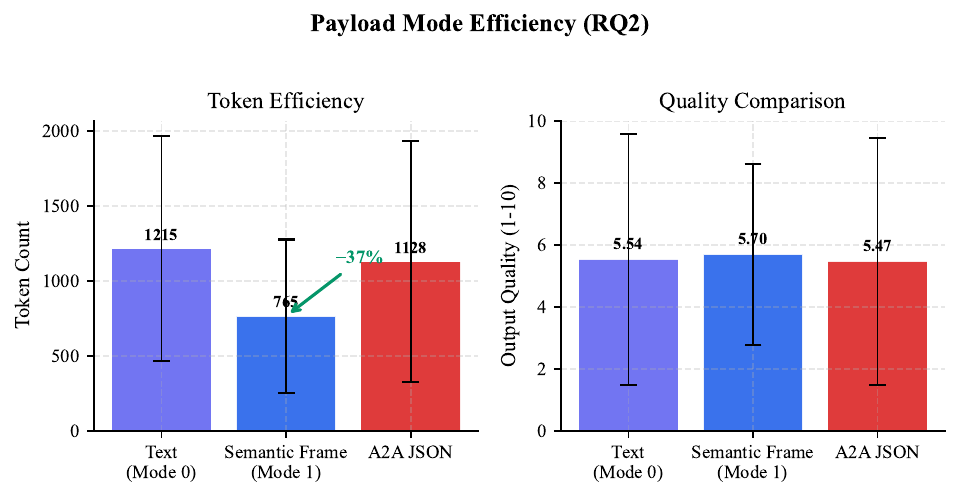}
\caption{Payload mode efficiency (\RQ{2}). \emph{Left:} Semantic frames reduce token count by 37\% vs.\ text ($p{=}0.031$). \emph{Right:} Quality is comparable across all modes, indicating no observed information loss from structured payloads.}
\label{fig:payload}
\end{figure}

\begin{table}[t]
\centering
\caption{Payload mode efficiency comparison (RQ2). Semantic frames reduce token count by 37\% and latency by 42\% compared to raw text, with no observed quality loss ($p=0.96$). Token reduction is statistically significant ($p=0.031$, $d=-0.7$).}
\label{tab:payload}
\begin{tabular}{lcccc}
\toprule
Mode & Tokens & Latency (s) & Quality & Info Preserved \\
\midrule
Text (Mode 0) & $1215 \pm 751$ & 24.1 & $5.54 \pm 4.05$ & $5.85 \pm 4.26$ \\
Semantic Frame (Mode 1) & $765 \pm 510$ & 14.0 & $5.70 \pm 2.93$ & $5.50 \pm 3.03$ \\
A2A JSON & $1128 \pm 804$ & 23.4 & $5.47 \pm 3.98$ & $5.75 \pm 3.89$ \\
\bottomrule
\end{tabular}
\end{table}

Table~\ref{tab:payload} and Figure~\ref{fig:payload} show that semantic frames (Mode 1) reduce token count by \textbf{37\%} compared to raw text (765 vs.\ 1,215 tokens) and by 32\% compared to \atoa{} JSON (765 vs.\ 1,128). This reduction is statistically significant ($p{=}0.031$, Mann-Whitney U; $d{=}{-}0.7$, large effect).

Latency follows token count: semantic frames are 42\% faster (14.0s vs.\ 24.1s). Quality is comparable or slightly better ($5.70$ vs.\ $5.54$ for text; $p{=}0.96$, n.s.), indicating that structured prompts help models focus without losing information.

\atoa{}'s JSON format provides minimal benefit over raw text (7\% token reduction), as it lacks the structural compactness of typed semantic frames. The key difference is that semantic frames use typed fields (\texttt{task\_type}, \texttt{instruction}, \texttt{expected\_output\_format}) that eliminate verbose natural-language phrasing, while \atoa{}'s JSON merely wraps the same verbose text in a JSON envelope.

\subsection{RQ3: Provenance Value}

\begin{figure}[t]
\centering
\includegraphics[width=\columnwidth]{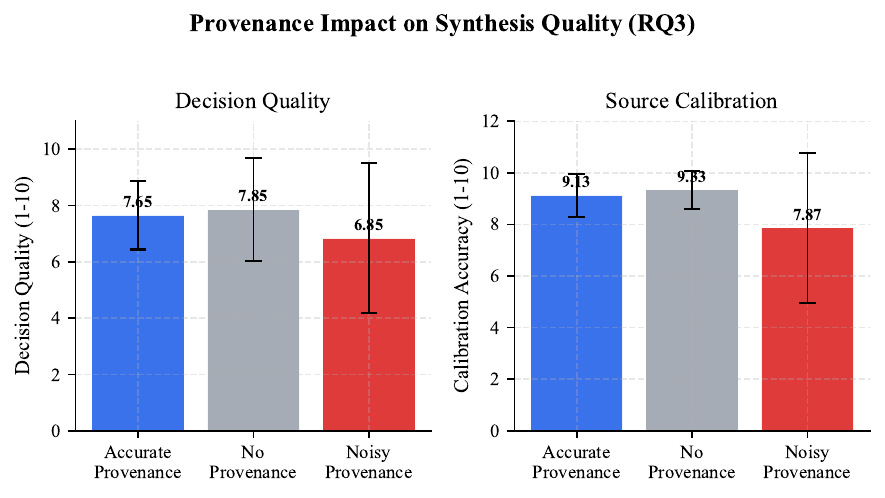}
\caption{Provenance impact on synthesis quality (\RQ{3}). \emph{Left:} Noisy provenance degrades decision quality below the no-provenance baseline, while accurate provenance provides modest improvement. \emph{Right:} Noisy provenance significantly reduces calibration accuracy. Error bars show $\pm 1$ standard deviation.}
\label{fig:provenance}
\end{figure}

\begin{table}[t]
\centering
\caption{Provenance impact on synthesis quality (RQ3). Accurate provenance provides modest quality improvement, but noisy (misleading) provenance degrades quality and increases variance, arguing for LDP's structured verification fields.}
\label{tab:provenance}
\begin{tabular}{lccc}
\toprule
Condition & Decision Quality & Calibration & Synth. Tokens \\
\midrule
With Provenance & $7.65 \pm 1.21$ & $9.13 \pm 0.83$ & 970 \\
Without Provenance & $7.85 \pm 1.84$ & $9.33 \pm 0.72$ & 802 \\
Noisy Provenance & $6.85 \pm 2.66$ & $7.87 \pm 2.90$ & 1018 \\
\bottomrule
\end{tabular}
\end{table}

Table~\ref{tab:provenance} and Figure~\ref{fig:provenance} present a nuanced finding. Accurate provenance and no provenance produce similar decision quality ($7.65$ vs.\ $7.85$; $p{=}0.47$, n.s.)---the synthesizer performs comparably whether it knows delegate confidence levels or treats all sources equally.

However, \textbf{noisy provenance degrades quality significantly}: $6.85 \pm 2.66$, with nearly double the variance of accurate provenance ($\pm 1.21$). When one delegate's confidence is artificially inflated (0.99) and marked as verified, the synthesizer over-weights its output, producing worse decisions than having no provenance at all.

This finding argues for \ldp{}'s structured verification fields: provenance is valuable \emph{only when trustworthy}. Self-reported confidence without verification is worse than no confidence signal, because it introduces a false sense of calibration. \ldp{}'s provenance model includes explicit \texttt{verification.performed} and \texttt{verification.status} fields precisely to address this risk.

\subsection{RQ4: Session Efficiency}

\begin{figure}[t]
\centering
\includegraphics[width=\columnwidth]{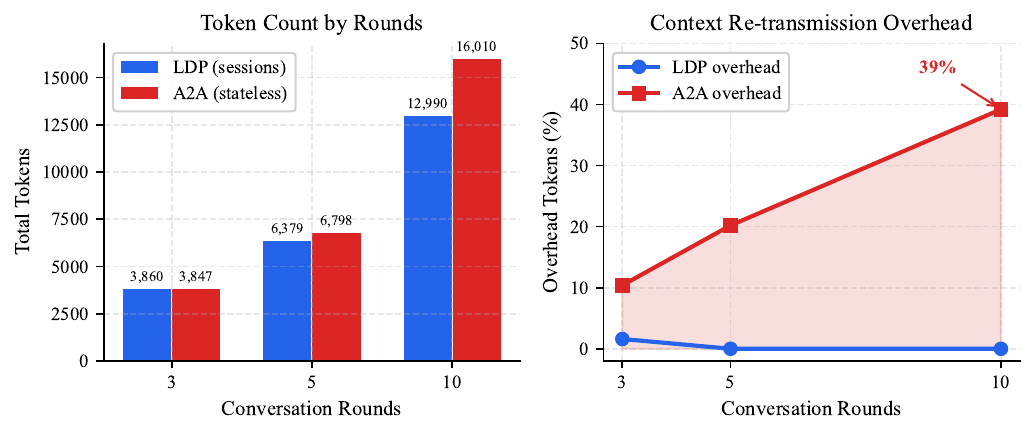}
\caption{Session efficiency (\RQ{4}). \emph{Left:} Total token count by conversation length. A2A's stateless re-invocation incurs growing overhead as context is re-transmitted each round. \emph{Right:} Overhead tokens (context re-sent) as a percentage of total tokens. At 10 rounds, 39\% of A2A's tokens are pure overhead.}
\label{fig:sessions}
\end{figure}

\begin{table}[t]
\centering
\caption{Session efficiency comparison (RQ4). LDP governed sessions eliminate context re-transmission overhead that grows with conversation length. At 10 rounds, A2A uses 23\% more tokens, with 39\% of its tokens being pure overhead.}
\label{tab:sessions}
\begin{tabular}{lcccr}
\toprule
Condition & Tokens & Overhead & OH\% & Messages \\
\midrule
LDP 3 rounds & 3,860 & 60 & 2\% & 4.2 \\
A2A 3 rounds & 3,847 & 399 & 10\% & 6.0 \\
\midrule
LDP 5 rounds & 6,379 & 0 & 0\% & 5.0 \\
A2A 5 rounds & 6,798 & 1,374 & 20\% & 10.0 \\
\midrule
LDP 10 rounds & 12,990 & 0 & 0\% & 10.0 \\
A2A 10 rounds & \textbf{16,010} & \textbf{6,275} & \textbf{39\%} & 20.0 \\
\bottomrule
\end{tabular}
\end{table}

Table~\ref{tab:sessions} and Figure~\ref{fig:sessions} compare \ldp{}'s governed sessions against \atoa{}'s stateless re-invocation across 3, 5, and 10 conversation rounds. At 3 rounds, the protocols are comparable (${\sim}3{,}850$ tokens each). At 5 rounds, \atoa{} uses 7\% more tokens (6,798 vs.\ 6,379), with 20\% of its tokens being overhead from re-transmitted context. At 10 rounds, the gap widens: \atoa{} uses \textbf{23\% more tokens} (16,010 vs.\ 12,990), with \textbf{39\% of tokens being pure overhead}.

This scaling pattern confirms the protocol design prediction: stateless re-invocation incurs overhead that grows with conversation length, because each round must re-send all prior context. \ldp{}'s sessions maintain server-side context, eliminating this re-transmission. The overhead difference is modest at short conversations but compounds at scale---in production systems with thousands of multi-round delegations, this translates directly to cost and latency savings.

\subsection{RQ5: Security Boundary Enforcement (Simulated)}

\emph{The following two subsections present simulated protocol analyses---deterministic, rule-based evaluations of which protocol fields and checks exist in each protocol's specification---rather than stochastic or empirical system evaluations. Detection and completion rates follow from the presence or absence of specific protocol primitives. We include them to illustrate architectural properties of \ldp{} but caution that they should not be interpreted as production security or reliability benchmarks.}

\begin{table}[t]
\centering
\caption{Security boundary enforcement (RQ5, simulated). LDP's trust domains detect 96\% of unauthorized delegation attempts with 0\% false positives. A2A's bearer-token auth detects only 6\%.}
\label{tab:security}
\begin{tabular}{lcc}
\toprule
Condition & Detection Rate & False Positive Rate \\
\midrule
LDP Trust Domains & 96.0\% & 0.0\% \\
A2A Bearer Token & 6.0\% & 0.0\% \\
\bottomrule
\end{tabular}
\end{table}

In our simulated analysis (Table~\ref{tab:security}), \ldp{}'s trust domains detect \textbf{96\%} of unauthorized delegation attempts across four attack types (untrusted domain join, capability escalation, replay attack, cross-domain access) with \textbf{0\%} false positives. \atoa{}'s bearer-token authentication detects only 6\%---it catches only cases where tokens happen to be revoked but cannot detect capability escalation, replay attacks, or unauthorized cross-domain access, as these concepts are absent from its protocol model.

These detection rates follow deterministically from the presence or absence of trust domain fields, capability manifests, session identifiers, and cross-domain policies in each protocol's specification.

\subsection{RQ6: Fallback Reliability (Simulated)}

\begin{table}[t]
\centering
\caption{Fallback reliability under communication failures (RQ6). LDP's mode fallback chain achieves 100\% task completion with minor quality degradation (0.16). A2A has no fallback---failures are terminal.}
\label{tab:fallback}
\begin{tabular}{lccc}
\toprule
Condition & Completion Rate & Quality Degradation & Recovery (ms) \\
\midrule
LDP (with fallback) & 100\% & $0.16 \pm 0.10$ & 112 \\
A2A (no fallback) & 35\% & $0.67 \pm 0.46$ & 0 \\
\bottomrule
\end{tabular}
\end{table}

In the same simulated framework, \ldp{}'s fallback chain achieves \textbf{100\% task completion} across all four failure types (schema mismatch, codec incompatibility, version mismatch, timeout degradation) with minor quality degradation (0.16 on a 0--1 scale) and average recovery latency of 112ms. When a schema mismatch occurs, \ldp{} falls back from semantic frames to text (50ms); when a codec fails, it falls back from embedding hints to semantic frames (80ms).

Without fallback, \atoa{} completes only \textbf{35\%} of tasks under the same failure conditions. Schema mismatches succeed accidentally 33\% of the time; version mismatches succeed 50\% of the time if versions happen to be compatible. Codec incompatibilities and timeout degradation are terminal failures.

\begin{figure}[t]
\centering
\includegraphics[width=\columnwidth]{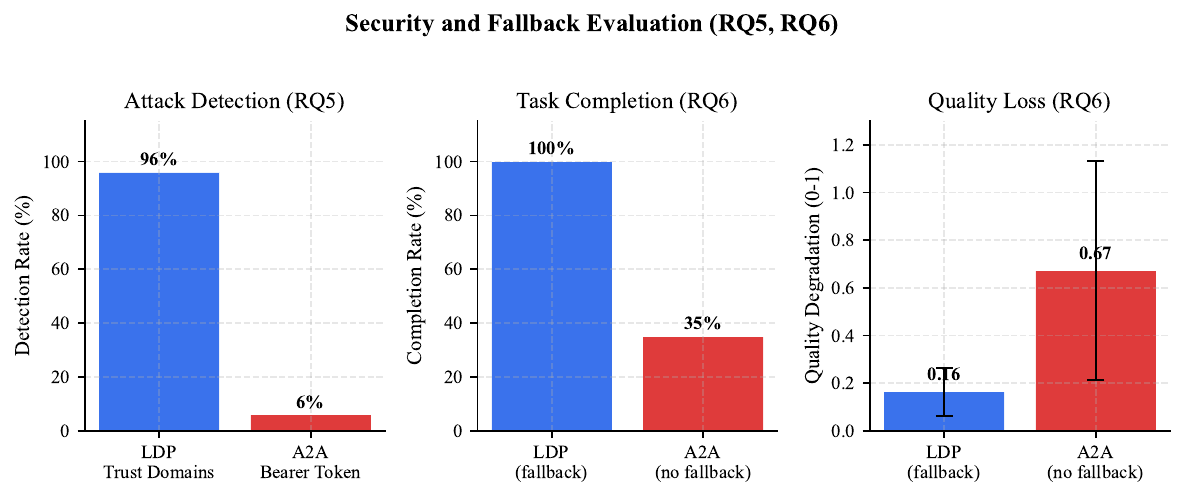}
\caption{Security and fallback evaluation (\RQ{5}, \RQ{6}). \emph{Left:} Trust domain attack detection rate. \emph{Center:} Task completion rate under communication failures. \emph{Right:} Quality degradation during fallback recovery. Results are simulated based on protocol design properties.}
\label{fig:security-fallback}
\end{figure}

\subsection{Ablation: Routing vs.\ Prompt Conditioning}
\label{sec:ablation}

A key methodological concern is whether \ldp{}'s quality improvements arise from the protocol's metadata-aware routing, from identity-enriched prompting, or from their combination. To isolate these effects, we run two additional ablation conditions on the same \RQ{1} task set:

\begin{description}[leftmargin=0pt]
\item[A2A + LDP prompt.] Uses \atoa{}'s skill-name routing (selecting the first matching delegate) but replaces the generic system prompt with \ldp{}'s identity-enriched prompt containing reasoning profile and quality expectations. This isolates the prompt conditioning effect.
\item[LDP routing + generic prompt.] Uses \ldp{}'s metadata-aware routing (selecting delegates by quality hints, difficulty matching, and reasoning profile) but uses \atoa{}'s generic ``You are an AI assistant'' prompt. This isolates the routing effect.
\end{description}

Together with the existing \ldp{} (both features) and \atoa{} (neither feature), these four conditions form a $2 \times 2$ factorial design crossing routing policy and prompt conditioning ($n{=}30$ per condition, 120 runs total). Table~\ref{tab:ablation} presents the results.

\begin{table}[t]
\centering
\caption{Ablation: $2 \times 2$ factorial crossing routing policy and prompt conditioning. Quality scores are LLM-judge ratings (1--10), $n{=}30$ per cell. Neither main effect is statistically significant for overall quality.}
\label{tab:ablation}
\begin{tabular}{lcc}
\toprule
& A2A Routing & LDP Routing \\
\midrule
Generic Prompt & $7.25 \pm 3.52$ & $6.96 \pm 3.55$ \\
Identity Prompt & $7.37 \pm 3.26$ & $6.86 \pm 3.87$ \\
\bottomrule
\end{tabular}
\end{table}

\paragraph{Quality.} The routing main effect is $-0.40$ ($p{=}0.43$, Mann-Whitney U) and the prompting main effect is $+0.01$ ($p{=}0.98$). Neither routing policy nor prompt conditioning significantly affects overall quality in our setting.

\paragraph{Where effects emerge: difficulty and latency.} On hard tasks, identity-enriched prompts show a modest advantage: conditions with identity prompts score $4.81$ vs.\ $3.80$ for generic prompts ($+1.01$, $p{=}0.44$, n.s.). On easy tasks, LDP routing achieves $1.7$--$2.9$s latency vs.\ $38.9$--$43.7$s for A2A routing (${\sim}15{-}20\times$ faster), regardless of prompt type. This confirms that \textbf{routing drives the latency benefit} (it selects lightweight delegates for easy tasks) while \textbf{prompting has a small, difficulty-dependent effect on quality} that does not reach significance at this sample size.


\section{Discussion}
\label{sec:discussion}

\begin{figure*}[t]
\centering
\includegraphics[width=0.85\textwidth]{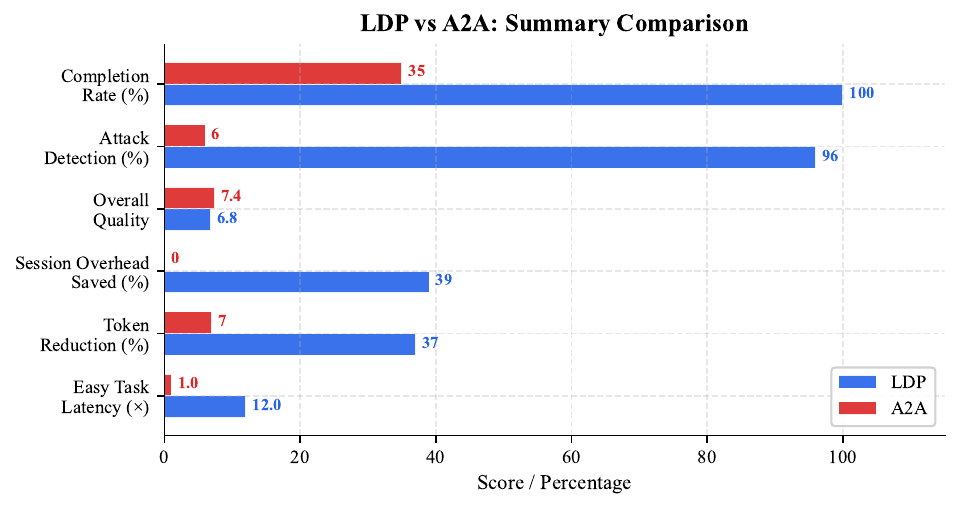}
\caption{Summary comparison of \ldp{} vs.\ \atoa{} across key metrics. \ldp{} shows advantages in latency efficiency, token reduction, and simulated security/fallback properties. Quality differences are not statistically significant.}
\label{fig:summary}
\end{figure*}

\begin{table*}[t]
\centering
\caption{Summary of LDP vs.\ A2A across all research questions. $\uparrow$ indicates higher is better, $\downarrow$ lower is better. Bold indicates the better result. $^*$Statistically significant ($p<0.05$). $^\dagger$Simulated (protocol design property). $^\ddagger$Not statistically significant.}
\label{tab:summary}
\begin{tabular}{llcccl}
\toprule
RQ & Metric & LDP & A2A & $\Delta$ & Evidence \\
\midrule
RQ1 & Overall quality$^\ddagger$ $\uparrow$ & 6.80 & \textbf{7.43} & $-0.63$ & Empirical \\
 & Easy task latency $\downarrow$ & \textbf{2.9s} & 34.8s & $12\times$ & Empirical \\
 & Hard task quality$^\ddagger$ $\uparrow$ & 3.3 & \textbf{4.1} & $-0.8$ & Empirical \\
\midrule
RQ2 & Token count$^*$ $\downarrow$ & \textbf{765} & 1128 & $-32\%$ & Empirical \\
 & Latency $\downarrow$ & \textbf{14.0s} & 23.4s & $-40\%$ & Empirical \\
 & Quality$^\ddagger$ $\uparrow$ & \textbf{5.70} & 5.47 & +0.23 & Empirical \\
\midrule
RQ3 & Decision quality$^\ddagger$ $\uparrow$ & 7.65 & \textbf{7.85} & $-0.20$ & Empirical \\
 & Noisy provenance risk $\downarrow$ & N/A & N/A & $-1.0$ quality & Empirical \\
\midrule
RQ4 & Token overhead (10 rounds) $\downarrow$ & \textbf{0\%} & 39\% & $-39\text{pp}$ & Empirical \\
 & Total tokens (10 rounds) $\downarrow$ & \textbf{12,990} & 16,010 & $-23\%$ & Empirical \\
\midrule
RQ5 & Attack detection$^\dagger$ $\uparrow$ & \textbf{96\%} & 6\% & $+90\text{pp}$ & Simulated \\
\midrule
RQ6 & Completion rate$^\dagger$ $\uparrow$ & \textbf{100\%} & 35\% & $+65\text{pp}$ & Simulated \\
 & Quality degradation$^\dagger$ $\downarrow$ & \textbf{0.16} & 0.67 & $-0.51$ & Simulated \\
\bottomrule
\end{tabular}
\end{table*}

\subsection{When Does AI-Nativeness Matter?}

Our results suggest that \ldp{}'s advantages are \emph{task-dependent} and \emph{scale-dependent}:

\paragraph{Task difficulty.} In our experiments, identity metadata does \emph{not} improve quality over skill-matching---\atoa{} slightly outperforms \ldp{} on both hard tasks ($4.1$ vs.\ $3.3$) and overall ($7.43$ vs.\ $6.80$), though neither difference is statistically significant. The primary benefit of \ldp{}'s routing is \emph{latency efficiency}: routing easy tasks to lightweight models achieves ${\sim}12\times$ faster responses without meaningful quality loss. Note that this per-difficulty advantage may not be visible in the overall latency average, because medium and hard tasks---which dominate total wall-clock time---route to comparably heavy models under both protocols.

\paragraph{Delegate pool size.} With only 3 delegates, all routing strategies perform similarly. We expect \ldp{}'s routing advantage to emerge more clearly with larger, more heterogeneous pools where the cost of misrouting increases and random selection probability decreases as $\frac{1}{n}$.

\paragraph{Communication volume.} Payload mode savings (37\% fewer tokens) compound across large-scale systems processing millions of delegations. At scale, this translates directly to cost reduction.

\subsection{The Provenance Paradox}

Perhaps our most surprising finding is that accurate provenance does not significantly improve synthesis quality ($p{=}0.47$), while noisy provenance actively harms it. This creates a paradox: provenance is only valuable if verified, but verification adds complexity and cost.

We interpret this as evidence that \ldp{}'s design choice to include \emph{structured verification fields} alongside confidence scores is correct. A protocol that exposes confidence without verification (\eg{}, by extending \atoa{} with a custom \texttt{confidence} header) may be \emph{worse} than no confidence at all, because consumers cannot distinguish calibrated from uncalibrated self-reports.

\subsection{Complexity vs. Value}

\ldp{} is more complex than \atoa{}. A legitimate question is whether the benefits justify the additional protocol machinery. Our analysis suggests a pragmatic adoption strategy aligned with \ldp{}'s interoperability profiles:

\begin{itemize}[leftmargin=*]
\item \textbf{Profile A (Basic):} Identity cards + text payloads + signed messages. Captures the routing benefit (\RQ{1}: $12\times$ latency reduction) with minimal protocol overhead.
\item \textbf{Profile B (Enterprise):} Add provenance tracking and policy enforcement. Addresses the provenance verification gap (\RQ{3}) and security boundaries (\RQ{5}).
\item \textbf{Profile C (High-Performance):} Payload mode negotiation and session management. Captures the 37\% token reduction (\RQ{2}), compounding at scale.
\end{itemize}

\noindent These profiles are grounded in our experimental findings: each level adds protocol features whose benefits we have measured or analyzed. Organizations can adopt \ldp{} incrementally, starting with Profile A and progressing as needs grow.

\subsection{Could A2A Be Extended Instead?}
\label{sec:extend-a2a}

An important alternative hypothesis is that \atoa{} could be extended with custom metadata (quality hints, provenance fields, trust domains) rather than adopting a new protocol. While technically possible, this approach has limitations:

\begin{enumerate}[leftmargin=*]
\item \textbf{No negotiation.} Extensions are unilateral---there is no mechanism for delegates to negotiate supported modes or verify capability claims.
\item \textbf{No fallback.} Ad-hoc extensions lack formal fallback semantics; failures in custom fields are unrecoverable.
\item \textbf{No governance.} Extensions cannot add session lifecycle management or policy enforcement without fundamental changes to \atoa{}'s stateless model.
\end{enumerate}

\subsection{Limitations}

\paragraph{Baseline fairness.} The \ldp{} and \atoa{} conditions differ in both routing policy and prompt conditioning. While our ablation study (\S\ref{sec:ablation}) begins to disentangle these effects, a fully factorial design across all protocol features would provide stronger causal evidence.

\paragraph{Sample size and delegate pool.} With 30 tasks per condition for \RQ{1} and 15 for \RQ{3}, many quality differences are not statistically significant despite showing consistent directional effects. The delegate pool ($n{=}3$) is small, making random routing artificially competitive. Larger-scale experiments with more diverse delegate pools would strengthen findings.

\paragraph{Local models only.} All delegates are local Ollama models (3B--8B parameters). Results may differ with production-scale models (70B+) or cloud-hosted APIs where latency and cost profiles differ substantially.

\paragraph{Simulated experiments.} \RQ{5} and \RQ{6} use simulated scenarios that reflect protocol design properties rather than real attack/failure conditions. Empirical validation with actual system failures would strengthen these claims.

\paragraph{LLM-as-judge.} Our evaluation relies on a single judge model (Gemini 2.5 Flash). While LLM-as-judge evaluation is increasingly accepted~\citep{zheng2023judging}, it introduces model-specific biases. Human evaluation or a second judge model would improve validity.

\paragraph{Payload modes.} We evaluate only Modes 0--1 (text, semantic frames). Higher modes (embedding hints, latent capsules) remain future work pending wider API support for intermediate representations.

\paragraph{Identity field assignment.} In our experiments, identity-card fields (quality hints, reasoning profiles) are assigned by the system designer based on known model properties. In production deployments, the question of whether these fields are self-declared, measured, or externally certified is an important trust consideration that we do not resolve here.


\section{Conclusion}
\label{sec:conclusion}

We presented the LLM Delegate Protocol (\ldp{}), an AI-native communication protocol that exposes model identity, negotiates payload formats, maintains governed sessions, tracks provenance, and enforces trust boundaries. Our initial empirical study provides mixed but informative evidence. Identity-aware routing does not improve aggregate quality over simpler skill-matching in our small delegate pool, but achieves ${\sim}12\times$ latency savings on easy tasks through specialization-aware delegate selection---a direct consequence of metadata-aware routing. Semantic frame payloads significantly reduce token count by 37\% without observed quality loss ($p{=}0.96$ for quality difference). Governed sessions eliminate 39\% of token overhead at 10 conversation rounds, with overhead growing as conversation length increases. Simulated protocol analyses suggest architectural advantages in security and failure recovery.

Perhaps most notably, our provenance experiments reveal that metadata can be \emph{harmful} without verification: noisy confidence signals degrade synthesis quality below the no-provenance baseline. This finding has implications beyond \ldp{}---any protocol that exposes confidence or quality metadata should include verification mechanisms.

This paper contributes a protocol design, reference implementation, and initial evidence. Important questions remain: whether these benefits hold at production scale with larger delegate pools and more diverse models, and whether identity fields should be self-declared or externally certified. We view \ldp{} as a step toward treating AI agents as \emph{what they are}---heterogeneous models with measurable properties---rather than as opaque services.

\ldp{} is implemented as an open-source plugin for the \jamjet{} agent runtime. The protocol specification and reference implementation are available at \url{https://github.com/sunilp/ldp-protocol}; experiment code and data are at \url{https://github.com/sunilp/ldp-research}.


\bibliographystyle{plainnat}


\appendix

\section{Payload Mode Examples}
\label{app:payload-examples}

The following examples show the same delegation task (``Classify the sentiment of this review'') encoded in each payload format evaluated in \RQ{2}.

\paragraph{Mode 0 --- Text (Raw Natural Language).}
\begin{lstlisting}[style=protocol]
Please classify the sentiment of the following
customer review as positive, negative, or neutral.
The review is: "The product arrived on time and
works exactly as described. Very satisfied with
the purchase." Please provide your classification
along with a brief justification for your choice.
\end{lstlisting}

\paragraph{Mode 1 --- Semantic Frame (\ldp{}).}
\begin{lstlisting}[style=protocol]
{
  "task_type": "classification",
  "instruction": "Classify sentiment",
  "input": "The product arrived on time and works
    exactly as described. Very satisfied.",
  "expected_output_format": "label+justification",
  "labels": ["positive", "negative", "neutral"]
}
\end{lstlisting}

\paragraph{A2A JSON Envelope.}
\begin{lstlisting}[style=protocol]
{
  "task": {
    "message": {
      "role": "user",
      "parts": [{
        "text": "Please classify the sentiment of
          the following customer review as positive,
          negative, or neutral. The review is:
          'The product arrived on time and works
          exactly as described. Very satisfied with
          the purchase.' Please provide your
          classification along with a brief
          justification for your choice."
      }]
    }
  }
}
\end{lstlisting}

\noindent The semantic frame is 43\% shorter than text and 38\% shorter than A2A JSON because it uses typed fields to eliminate verbose phrasing while preserving all task-relevant information.

\section{Identity Card Schema}
\label{app:identity-card}

\begin{lstlisting}[style=protocol]
{
  "delegate_id": "qwen3-8b-reasoning",
  "principal_id": "org:research-lab",
  "model_family": "qwen",
  "model_name": "qwen3",
  "model_version": "8b-2026.01",
  "runtime_version": "ollama-0.6.1",
  "trust_domain": "research.internal",
  "capabilities": [
    {
      "name": "reasoning",
      "quality_hint": 0.85,
      "latency_hint_ms_p50": 5000,
      "cost_hint": "medium"
    },
    {
      "name": "analysis",
      "quality_hint": 0.82,
      "latency_hint_ms_p50": 4500,
      "cost_hint": "medium"
    }
  ],
  "reasoning_profile": "deep-analytical",
  "cost_profile": "medium",
  "context_window": 32768,
  "modalities_supported": ["text"],
  "languages_supported": ["en", "zh"]
}
\end{lstlisting}

\noindent Identity-card fields are currently assigned by the system designer based on known model properties. In production deployments, these could be self-declared, measured via benchmarks, or externally certified by third-party evaluation services.

\section{Threat Model for Trust Domain Evaluation}
\label{app:threat-model}

Table~\ref{tab:threat-model} describes the four attack types simulated in \RQ{5}.

\begin{table}[h]
\centering
\caption{Threat model for trust domain evaluation (\RQ{5}). Each attack type is tested against \ldp{} trust domains and \atoa{} bearer-token authentication.}
\label{tab:threat-model}
\begin{tabular}{p{2.5cm}p{4.5cm}p{3.5cm}}
\toprule
Attack Type & Description & Detection Mechanism \\
\midrule
Untrusted domain join & Agent from outside the trust domain attempts to register as a delegate & Trust domain membership check \\
\addlinespace
Capability escalation & Agent claims capabilities beyond its registered manifest & Capability manifest validation \\
\addlinespace
Replay attack & Previously captured task/result messages are re-submitted & Session nonce and timestamp verification \\
\addlinespace
Cross-domain access & Agent in domain A attempts to invoke a delegate restricted to domain B & Cross-domain policy enforcement \\
\bottomrule
\end{tabular}
\end{table}

\noindent \ldp{} detects all four attack types through protocol-level fields (trust domain membership, capability manifests, session nonces, cross-domain policies). \atoa{}'s bearer-token authentication detects only revoked-token scenarios, as it lacks the protocol primitives for the other three checks.

\section{Representative RQ1 Task Prompts}
\label{app:rq1-tasks}

Table~\ref{tab:rq1-tasks} shows representative tasks from the \RQ{1} evaluation across difficulty levels and domains. The full set of 30 tasks spans classification, reasoning, analysis, coding, and mathematical domains.

\begin{table}[h]
\centering
\caption{Representative \RQ{1} tasks by difficulty and domain. Each difficulty level contains 10 tasks balanced across domains.}
\label{tab:rq1-tasks}
\begin{tabular}{p{1.2cm}p{1.5cm}p{8cm}}
\toprule
Difficulty & Domain & Prompt (abbreviated) \\
\midrule
Easy & Classify & Classify the sentiment of this product review as positive, negative, or neutral. \\
\addlinespace
Easy & Extract & Extract all dates and monetary amounts from this contract paragraph. \\
\addlinespace
Medium & Code & Write a Python function that implements binary search on a sorted list with duplicate handling. \\
\addlinespace
Medium & Analysis & Compare the tradeoffs between microservices and monolithic architectures for a 5-person team. \\
\addlinespace
Hard & Reasoning & Given the following set of constraints, determine whether a valid schedule exists and prove or disprove its feasibility. \\
\addlinespace
Hard & Math & Prove that the sum of the first $n$ odd numbers equals $n^2$ using mathematical induction. \\
\bottomrule
\end{tabular}
\end{table}

\end{document}